\documentclass{article}

\usepackage{microtype}
\usepackage{graphicx}
\usepackage{subcaption}
\usepackage{booktabs} % for professional tables
\usepackage{soul}
\usepackage{hyperref}

\usepackage[preprint]{icml2026}

\usepackage{amsmath}
\usepackage{amssymb}
\usepackage{mathtools}
\usepackage{amsthm}

\usepackage[capitalize,noabbrev]{cleveref}

\theoremstyle{plain}

\theoremstyle{definition}

\theoremstyle{remark}

\usepackage{nicefrac}
\usepackage{xspace}
\usepackage[inline]{enumitem}
\usepackage[subpreambles=true]{standalone}
\usepackage{tikz}
\usepackage{mystyle}
\usepackage{multirow}
\usepackage{listings}
\newfloat{listing}{htbp}{lop}
\floatname{listing}{Listing}
\newcommand{\method}{Plan, Align, Commit, Think\xspace}
\newcommand{\abbrev}{PACT\xspace}

\usepackage[
    textsize=scriptsize,
]{todonotes}

\icmltitlerunning{When in Doubt, Plan It Out: Committed Small Language Model Deliberation for Reactive Reinforcement Learning}

\begin{document}

\twocolumn[
  \icmltitle{When in Doubt, Plan It Out: Committed Small Language Model Deliberation\\for Reactive Reinforcement Learning}

  % It is OKAY to include author information, even for blind submissions: the
  % style file will automatically remove it for you unless you've provided
  % the [accepted] option to the icml2026 package.

  % List of affiliations: The first argument should be a (short) identifier you
  % will use later to specify author affiliations Academic affiliations
  % should list Department, University, City, Region, Country Industry
  % affiliations should list Company, City, Region, Country

  % You can specify symbols, otherwise they are numbered in order. Ideally, you
  % should not use this facility. Affiliations will be numbered in order of
  % appearance and this is the preferred way.
  \icmlsetsymbol{equal}{*}

  \begin{icmlauthorlist}
    \icmlauthor{Nathan Gavenski}{equal,yyy}
    \icmlauthor{Juarez Monteiro}{equal,comp}
    \icmlauthor{Francisco Galuppo}{comp}
    \icmlauthor{Adriano Veloso}{comp}
    \icmlauthor{Odinaldo Rodrigues}{yyy}
  \end{icmlauthorlist}

  \icmlaffiliation{yyy}{Department of Informatics, King's College London, London, United Kingdom}
  \icmlaffiliation{comp}{Kunumi Institute, Belo Horizonte, Brazil}

  \icmlcorrespondingauthor{Nathan Gavenski}{nathan.schneider\_gavenski@kcl.ac.uk}
  \icmlcorrespondingauthor{Juarez Monteiro}{juarez@kunumi.com}

  % You may provide any keywords that you find helpful for describing your
  % paper; these are used to populate the "keywords" metadata in the PDF but
  % will not be shown in the document
  \icmlkeywords{Planning, Small Language Model, Reinforcement Learning, Generalisation}

  \vskip 0.3in
]

% this must go after the closing bracket ] following \twocolumn[ ...

% This command actually creates the footnote in the first column listing the
% affiliations and the copyright notice. The command takes one argument, which
% is text to display at the start of the footnote. The \icmlEqualContribution
% command is standard text for equal contribution. Remove it (just {}) if you
% do not need this facility.

% Use ONE of the following lines. DO NOT remove the command.
% If you have no special notice, KEEP empty braces:
% \printAffiliationsAndNotice{}  % no special notice (required even if empty)
% Or, if applicable, use the standard equal contribution text:
\printAffiliationsAndNotice{\icmlEqualContribution}

\begin{abstract}
Reinforcement Learning (RL) policies often degrade in unfamiliar environments because they lack explicit deliberation.
We propose \method (\abbrev), a hybrid architecture that combines a fast, reactive RL policy with a slow, deliberative Small Language Model (SLM) planner.
\abbrev invokes the SLM asynchronously to generate and validate candidate action plans.
Once a plan is verified through simulation as safe, feasible, and complete, it is executed directly, bypassing the RL policy without retraining or modifying it.
Evaluated on three FrozenLake configurations of increasing difficulty, \abbrev outperforms all baselines while relying on a $2$B-parameter SLM backbone, suggesting that deliberative planning and reactive execution are more powerful in concert than either is alone in these settings.
\end{abstract}

\section{Introduction}

Agentic systems are increasingly deployed in real-world applications where they must not only react to their immediate observations, but also plan sequences of actions to achieve long-horizon goals.
Although reinforcement learning (RL) has made significant progress in training capable agents, standard model-free RL policies remain fundamentally \emph{reactive}: they map states to actions without deliberation or lookahead, degrading systematically when deployed beyond training conditions.

This limitation has long been recognised:
\citeauthor{kahneman2012thinking}'s dual-process theory~\yrcite{kahneman2012thinking} distinguishes \emph{fast} thinking, automatic and reactive, from \emph{slow} thinking, deliberate and goal-directed.
Trained RL agents are natural fast thinkers: reliable within familiar territory, but incapable of the forward reasoning that novel situations demand.
A similar limitation occurs within symbolic AI paradigms. Belief-Desire-Intention (BDI) systems~\cite{bratman1999intention} introduced \emph{committed plans} to enable faster, reactive behaviour, although they still require manually engineered planning models and struggle to scale to real-world complexity.

Language models (LMs) offer a compelling path forward: pre-trained on vast corpora, they encode transferable knowledge and commonsense reasoning capabilities that suggest a natural role as high-level planning assistants for reactive agents.
However, LMs are known to struggle with multi-step planning on their own~\cite{ValmeekamKambhampati2025Frontier,armony2025far}, and large models are computationally expensive, making their direct use as planners impractical for low-latency deployment.
Small LMs (SLMs)~\cite{wang2024comprehensive} provide a more practical trade-off: computationally efficient yet capable enough to propose \emph{candidate action sequences} that, once verified, provide the structured guidance reactive agents cannot generate on their own.

We therefore propose \method (\abbrev), a hybrid architecture that combines a fast component, a reactive RL policy, with a slow component, a deliberative SLM planner, drawing on dual-process theory and BDI-style commitment.
\abbrev relies on the RL policy for efficient decision-making in familiar conditions, and invokes the SLM asynchronously when the agent's epistemic uncertainty exceeds a threshold, signalling that the current observation is outside its training distribution.
Once a plan is generated and verified as safe, the RL policy is bypassed: the agent commits to the plan until completion, revision, or failure.
This commitment mechanism, and the clean separation of reactive and deliberative control it affords, distinguishes \abbrev from approaches that treat LMs as step-level advisors~\cite{monteiro2026ask,saycan2022arxiv}.
Experiments on FrozenLake across three increasingly difficult configurations demonstrate that, without retraining, this hybrid architecture enables agents to handle situations that neither component can resolve on its own.

\section{Background}\label{sec:background}

Reinforcement learning is a problem in which an agent learns to maximise a reward signal by interacting with an environment.
More formally, the agent aims to learn a policy $\policy(a \mid s): \StateMDP \times \Actions \to [0, 1]$ that maps states $s \in \StateMDP$ to a distribution over actions $a \in \Actions$, by maximising the expected cumulative reward over time.
We formulate the problem in this work as a contextual Markov Decision Process (CMDP), which is a tuple $\mdp = \langle \StateMDP, \Actions, \Observations, \reward, \Transition, \Context, \Projection \rangle$, where $\Observations$ is the observation space, $\reward$ is the immediate reward function $\reward: \Observations \times \Actions \to \mathbb{R}$, $\Transition$ is the transition function $\Transition: \Observations \times \Actions \to \Observations$, $\Context$ is the context space, and $\Projection$ is the projection function $\Projection: \StateMDP \times \Context \to \Observations$ that maps states and contexts to observations~\cite{kirk2023survey}.
Note that in CMDPs, the agent observes the observation produced by $\Projection$, and the current context $c \in \Context$ remains the same in an episode.
In other words, the policy perceives only one context $c$ at a time in which all states $s \in \StateMDP$ are projected into it $\left\{ \Projection(s, c) \mid s \in \StateMDP \right\}$.
We do not assume that the agent has access to the context either via input or that it can be inferred from the observations, since two different contexts $c_1, c_2 \in \Context$ may map two different states $s_1, s_2 \in \StateMDP$ to the same observation $\Projection(s_1, c_1) = \Projection(s_2, c_2)$.
This framing makes the generalisation challenge precise: because the training and evaluation contexts are disjoint, a purely reactive policy $\policy(a \mid o)$ operating on observations alone cannot anticipate the environment's behaviour in novel contexts, which is the gap \abbrev is designed to close.

Finally, we use uncertainty estimation to determine when the RL policy is unsure about the next action.
%\otr{This is critical and should have been mentioned in the introduction. Charles Higgins' PhD spent considerable effort trying to come up with a solution for this.}
Commonly, uncertainty in neural networks can be split into two categories: aleatoric, which stems from the inherent randomness in the data, and epistemic, which arises from the model's lack of knowledge about the data~\cite{kendall2017what}.
Since aleatoric uncertainty is irreducible, we focus on epistemic.
To compute the epistemic uncertainty of an observation, we use the Monte Carlo Dropout method~\cite{kendall2017what}, which approximates Bayesian inference by performing multiple stochastic forward passes through the network with dropout enabled.
Thus, the epistemic uncertainty of an observation $o$, $\mathcal{U}(o)$, is defined as $\mathcal{U}(o) = \frac{1}{N} \sum_{i=1}^{N} \sum_{a} p_i(a \mid o) \log \frac{p_i(a \mid o)}{\bar{p}(a \mid o)}$,
where $p_i(a \mid o)$ is the action distribution from the $i$-th forward pass with dropout, and $\bar{p}(a\mid o) = \frac{1}{N} \sum_{i=1}^{N}p_i(a\mid o)$ is the mean distribution (we omit $\agent$ from $p_i^\agent$ for simplicity); this requires the policy to produce explicit per-action likelihoods.

\section{\method}

We propose \method (\abbrev), a hybrid architecture that combines a fast component, a reactive pre-trained RL policy, with a slow component, a deliberative SLM planner, drawing on dual-process theory and the recent advances in planning with LMs~\cite{ValmeekamKambhampati2025Frontier}.
\abbrev uses the RL policy for efficient local decision-making, while invoking the SLM asynchronously to generate candidate long-horizon plans. Because SLMs remain unreliable as standalone multi-step planners~\cite{monteiro2026ask}, \abbrev decomposes the planning process into three modules: plan generation, plan verification, and agent alignment; unlike step-level approaches such as ASK~\cite{monteiro2026ask}, which query the SLM independently at each uncertain step with no forward plan or commitment.
\abbrev further requires an approximate transition function for plan simulation and a task description sufficient for SLM prompting (see Appendix~\ref{ap:arch_and_prompts}).

%\abbrev relies on the RL agent for local, efficient decision-making and asynchronously invokes the SLM at the start of each episode to generate a candidate plan for the entire trajectory. However, SLMs are known to struggle with multi-step planning in isolation~\cite{monteiro2026ask}, making their direct use as a planner unreliable. Therefore, we break down the planning process into three steps: plan generation, plan verification, and agent alignment.

\begin{figure}
    \centering
    \begin{tikzpicture}[node distance=0.6cm]

    % ---------------------------------------------------------------
    % Env
    % ---------------------------------------------------------------
    \node [rectangle, rounded corners, draw=cb_green!80, fill=cb_green!15,
           text centered, align=center, font=\small] (env) {Env.};

    % ---------------------------------------------------------------
    % Slow Agent Modules  (top row)
    % ---------------------------------------------------------------
    \node [module, right=2.2cm of env, yshift=1.4cm] (planner) {Planner};
    \node [decision, right=0.55cm of planner] (oracle) {done?};
    \node [module, right=1.2cm of oracle] (verification) {Verification};
    \node [decision, right=0.55cm of verification] (safe) {safe?};
    \node [module, right=1.2cm of safe] (alignment) {Alignment};
    % \node [state, accepting, inner sep=1pt, minimum size=0.3cm, font=\normalsize, right=1.2cm of alignment] (end_slow) {$\plan$};

    % Slow arrows
    \draw [arrow] (planner) -- (oracle);
    \draw [arrow] (oracle) -- 
        node[above,font=\normalsize, pos=0.35]{yes} 
        node[below,font=\small, pos=0.4, align=center]{\textit{candidate}\\\textit{plan}} 
        (verification);
    \draw [arrow] (verification) -- (safe);
    \draw [arrow] (safe) -- 
        node[above,font=\normalsize, pos=0.35]{yes}
        node[below,font=\small, pos=0.4, align=center]{\textit{verified}\\\textit{plan}}
        (alignment);
    
    % Loop back: not done
    \draw [arrow] (oracle.south) -- ++(0,-0.5) -| 
        node[above,font=\normalsize,pos=0.25]{no}
        node[below,font=\small,pos=0.25]{\textit{next action}} 
        (planner.south);
    \draw [arrow] (safe.north) -- ++(0,0.35)
        -| node[above,font=\normalsize,pos=0.35]{no} (planner.north);

    % o_1 from env to planner
    \draw [arrow] (env.east) -- 
        node[xshift=6pt, above,font=\normalsize,pos=0.25]{$\Projection(s_t, c)$}
        node[xshift=6pt, below,font=\normalsize,pos=0.25]{$s_{t+1}$}
        ++(1.25,0) |-
        node[xshift=6pt, above,font=\normalsize,pos=0.15]{$o$}
        (planner.west);

    % Update alignment
    \draw [arrow] (env.north) -- +(0, 3) -| 
        node[xshift=6pt, below, font=\normalsize,pos=0.25]{\emph{update belief} $(o_t, a_t)$}
        (alignment.north);

    % ---------------------------------------------------------------
    % Fast Agent  (bottom row)
    % ---------------------------------------------------------------
    \node [decision, right=3cm of env, yshift=-1.9cm] (d_plan) {plan\\ready?};
    \node [decision, right=1.2cm of d_plan] (d_goal) {$o_t{\in}\Goals$?};

    \node [module, right=1.2cm of d_goal, minimum width=0.9cm] (ppo) {PPO $\pi_\theta$};
    \node [decision, right=1.2cm of ppo]   (d_unc)  {$\mathcal{U}(o_t){\geq}\tau$?};

    % obs to ppo
    \draw [arrow] (env.east) -- ++(1.25,0)
        |- node[xshift=6pt, below, font=\normalsize, pos=0.25]{$o_t$} (d_plan.west);

    % PPO → decisions
    \draw [arrow] (ppo) -- (d_unc);
    \draw [arrow] (d_goal) -- node[above,font=\normalsize]{no} (ppo);
    \draw [arrow] (d_unc.south) -- ++(0,-1) -| 
        node[above,font=\normalsize, pos=0.1]{no, $a_t \gets \pi_\theta(o_t)$} 
        node[right,font=\normalsize, pos=0.7]{$a_t$} 
        (env.south);

    % (i) goal reached → end
    \node [state, accepting, inner sep=1pt, minimum size=0.3cm, font=\normalsize, below=0.5cm of d_goal] (end) {$g$};
    \draw [arrow] (d_goal.south)
        -- node[right,font=\normalsize]{yes} (end);

    % (ii) uncertainty → wait for plan (arrow up toward slow agent)
    \node [circle, fill=black, minimum size=0.25cm, above=0.4cm of d_unc, label={[font=\normalsize]above:wait}] (wait) {};
    \draw [arrow] (d_unc.north) -- node[right,font=\normalsize]{yes} (wait);
    \draw [darrow, -] (wait.west) -- ++(0.2,0) -| ([yshift=14pt]d_plan.north);

    % (iii) plan ready → execute plan → loops to PPO
    % \node [font=\normalsize, below=0.55cm of d_plan, align=center] (exec) {execute\\plan};
    \draw [arrow, -] (d_plan.south) -- node[right, font=\normalsize, align=left]{yes,\\$a_t \gets \plan[t]$} ++(0,-1);
    % plan execution loops back
    % \draw [arrow] (exec.south) -- ++(0,-0.3)
    %     -| ($(ppo.south)+(0,-0.3)$) -- (ppo.south);

    % (iii) no plan yet → PPO acts (loop back)
    \draw [arrow] (d_plan.east) -- node[above,font=\normalsize,pos=0.25]{no} (d_goal.west);

    % ---------------------------------------------------------------
    % Alignment → d_plan  (dashed: plan committed)
    % ---------------------------------------------------------------
    \draw [arrow] (alignment.east) -- 
        ++(0.5, 0) |-
        ++(0, -1.65) -|
        node[above, font=\small, pos=0.25, align=center, yshift=-2pt]{\textit{aligned plan}}
        (d_plan.north);

    % ---------------------------------------------------------------
    % Background boxes
    % ---------------------------------------------------------------
    \begin{scope}[on background layer]
        \node [%
            agent,
            fit=(planner)(verification)(alignment)(oracle),
            inner sep=8pt,
            label={[anchor=north, yshift=-1pt, inner sep=2pt, font=\normalsize]north:{Slow Component}},
            draw=cb_blue!80, fill=cb_blue!15,
            minimum height=70pt
        ] (slow) {};

        \node [%
            agent,
            fit=(ppo)(d_goal)(d_unc)(d_plan)(end)(wait),
            inner sep=8pt,
            label={[anchor=north, yshift=-1pt, inner sep=2pt, font=\normalsize]north:{Fast Component}},
            draw=cb_orange!80, fill=cb_orange!15,
            yshift=-0cm,
            minimum height=90pt
        ] (fast) {};
    \end{scope}

\end{tikzpicture}
    \caption{\method architecture.}
    \label{fig:method}
    \vskip -0.5cm
\end{figure}

\textbf{Process Overview}
Given an observation $o$, \abbrev routes it simultaneously to both components: the slow component initiates plan generation, verification, and alignment, while the fast component acts via the RL policy.
\abbrev computes the uncertainty of the policy for every new observation $o$, $\mathcal{U}(o)$. If it is smaller than the threshold $\tau$, the agent continues to apply the fast component until one of three conditions are met:
\begin{enumerate*}[label=(\roman*)]
    \item the uncertainty of the current observation $o_t$ goes above the threshold $\tau$;
    \item the slow component successfully generates a plan that is both verified and aligned with the agent's current trajectory; or
    \item the goal $g$ is reached, thus, requiring no further action.
\end{enumerate*}
If the uncertainty of the RL policy exceeds the uncertainty threshold for any observation, \abbrev waits for $t$ time units for the plan generation, verification, and alignment completion. If the timeout expires, the RL component is reactivated.
However, if the plan becomes ready before the agent reaches the goal, \abbrev bypasses the fast component and executes it; otherwise the process restarts.
We note that \abbrev updates the alignment module's beliefs at each agent action.
Thus, once an aligned plan is generated, no further alignment is necessary unless the agent encounters unexpected transitions (i.e., stochastic transition functions); plans generated from earlier observations are reconciled with the current position by the alignment module before committing.

\textbf{Plan Generation Module}
This module generates candidate plans $\rho$ with up to $n$ actions.
Initially, the SLM is given a description of the task and the episode's first observation $o_1$, and the slow component asks it for the action that takes it closest to the goal.
The remaining actions $a_2, a_3, \ldots$ are generated by successively prompting the SLM until an observation indicates that the goal has been reached or the last action $a_n$ is generated.
Due to the SLM's limited capabilities, it is important to include as much task information as possible in the prompts used for each action generation.
For simplicity, we simulate action execution using a hand-crafted transition function and update each prompt with the simulated observations.
More sophisticated action simulation methods could be used instead, e.g., by learning a dynamics model from data~\cite{gavenski2026towards}.
Crucially, the transition function does not need to be perfect: a heuristic approximation that captures the dominant dynamics suffices in this setting, since any residual mismatch between the simulated and actual environment is handled by the alignment module's replanning mechanism.

\textbf{Plan Verification Module}
Candidate plans are simulated to assess their feasibility, safety and completeness.
Firstly, the slow component checks that the plan is deployable given its action space and capabilities.
Secondly, the slow component checks for any violation of safety constraints.
If any constraint is violated, the slow component prompts the SLM to generate a new candidate plan that avoids the violation, and the process is repeated until both modules agree on a safe plan.
Finally, although \abbrev works iteratively and can handle incomplete plans, candidate plans are checked for completeness to avoid unnecessary iterations, since the alignment process may be costly if the agent and the candidate plan differ significantly.

\textbf{Agent Alignment Module}
\abbrev uses the SLM as an asynchronous planner, and the RL policy as a reactive executor.
The alignment module bridges the gap between the idealised environment assumed during plan generation and verification and the agent's actual current state.
If a discrepancy is found, \abbrev iteratively prompts the SLM, one action at a time, to guide the agent toward the closest reachable waypoint in the verified plan.
Crucially, this process is \textit{goal-directed}: the alignment module commits to a specific re-entry point in the original plan, and terminates as soon as the agent reaches it, after which committed execution of the verified plan resumes.
This distinguishes the alignment module from step-level advisory approaches, which have no target waypoint and no termination criterion tied to plan re-entry.
After alignment, \abbrev executes the remaining verified plan by bypassing the fast component and directly issuing actions in the environment.

\section{Experimental Analysis}

\begin{table*}
  \scriptsize
  \centering
  \caption{Results across environments for \abbrev and baselines.}
  \label{tab:results}
  \begin{tabular*}{\textwidth}{@{\hspace{2pt}}l@{\extracolsep{\fill}} rrr rrr rrr@{\hspace{2pt}}}
    \toprule%\noalign{\vspace{-1pt}}
    & \multicolumn{3}{c}{$6 \times 6$} & \multicolumn{3}{c}{$6 \times 6$ (With Slipperiness)} & \multicolumn{3}{c}{$8 \times 8$} \\
    \cmidrule{2-4}\cmidrule{5-7}\cmidrule{8-10}
    Methods & Reward & Length & LM Usage & Reward & Length & LM Usage & Reward & Length & LM Usage \\
    \cmidrule{1-1}\cmidrule{2-4}\cmidrule{5-7}\cmidrule{8-10}
    \abbrev & $\mathbf{0.98 \pm 0.14}$ & $10.0 \pm 0.9$ & $27.9\%$ & $\mathbf{0.93 \pm 0.26}$ & $9.5 \pm 1.9$ & $58.4\%$ & $\mathbf{1.00 \pm 0.00}$ & $15.5 \pm 2.2$ & $81.2\%$ \\
    \cmidrule{1-1}\cmidrule{2-4}\cmidrule{5-7}\cmidrule{8-10}
    PPO     & $0.93 \pm 0.26$ & $9.5 \pm 1.9$  & --        & $0.64 \pm 0.48$ & $9.6 \pm 4.2$  & --        & $0.79 \pm 0.41$ & $12.4 \pm 3.4$  & -- \\
    \cmidrule{1-1}\cmidrule{2-4}\cmidrule{5-7}\cmidrule{8-10}
    SLM     & $0.47 \pm 0.50$ & $81.2 \pm 66.7$ & $100\%$  & $0.47 \pm 0.50$ & $15.5 \pm 17.7$ & $100\%$  & $0.42 \pm 0.49$ & $35.4 \pm 37.6$ & $100\%$ \\
    \cmidrule{1-1}\cmidrule{2-4}\cmidrule{5-7}\cmidrule{8-10}
    SAYCAN  & $0.88 \pm 0.33$ & $9.3 \pm 1.9$  & $60.1\%$  & $0.53 \pm 0.50$ & $9.1 \pm 4.1$  & $59.7\%$  & $0.62 \pm 0.49$ & $11.6 \pm 3.6$  & $73.1\%$ \\
    \cmidrule{1-1}\cmidrule{2-4}\cmidrule{5-7}\cmidrule{8-10}
    ASK     & $0.89 \pm 0.31$ & $14.4 \pm 21.8$ & $27.1\%$ & $0.63 \pm 0.48$ & $10.7 \pm 5.3$  & $26.7\%$ & $0.74 \pm 0.44$ & $13.4 \pm 3.8$  & $28.7\%$ \\
    \bottomrule
  \end{tabular*}
  \vskip -5pt
\end{table*}

We evaluated \abbrev on FrozenLake~\cite{towers2024gymnasium} on three different configurations:
\begin{enumerate*}[label=\roman*)]
  \item a $6\times6$ map without slipperiness, for control over the episode termination;
  \item a $6\times6$ map with slipperiness, to test the agent's ability to handle stochasticity; and
  \item a $8\times8$ map, without slipperiness, to test the agent's ability to handle large state spaces with sparse rewards and long horizons.
\end{enumerate*}
For the `slow' component (SLM), we used a $2$B model from~\citet{qwen3.5} as a zero-shot SLM backbone.
The `fast' component used Proximal Policy Optimisation (PPO)~\cite{schulman2017proximal} trained on $100$ contexts not present in the evaluation and test splits.
We compare PACT, PPO, and the standalone SLM with SAYCAN~\cite{saycan2022arxiv} and ASK~\cite{monteiro2026ask}, which are state-of-the-art LM-RL approaches.
All baselines were fine-tuned in $100$ separate contexts.
Full implementation details are in Appendix~\ref{ap:arch_and_prompts}--\ref{ap:baselines}; $\tau$ is selected on a held-out validation set (Appendix~\ref{ap:threshold}).

Table~\ref{tab:results} reports reward, episode length, and LM usage for all agents across the three configurations.
\abbrev achieves the highest reward in every setting, from $0.98 \pm 0.14$ on the $6\times6$ map, to $0.93 \pm 0.26$ under stochasticity, to a perfect $1.00 \pm 0.00$ on the $8\!\times\!8$ map with zero variance, consistently outperforming the agents based on its individual components as well as the other LM--RL approaches.

The $6\!\times\!6$ slippery configuration isolates the effect of stochasticity on each method, with all baselines degrading substantially: PPO drops from $0.93$ to $0.64$, SAYCAN from $0.88$ to $0.53$, and ASK from $0.89$ to $0.63$.
\abbrev's performance, by contrast, drops only by $5\%$ (to a still respectable $0.93$), remaining the top-performing method by a wide margin.
Stochastic transitions frequently deviate the agent from its committed plan, leading to $2.85$ replans per episode on average and a corresponding increase in LM usage to $58.4\%$.
The most telling comparison is with SAYCAN, which consults the LM at a nearly identical rate ($59.7\%$) yet achieves only $0.53$ reward.
This gap highlights the effect of plan commitment: both methods query the LM with similar frequency, but \abbrev uses those queries to generate, verify, and commit to structured trajectories, while SAYCAN issues independent step-level proposals with no higher-level structure to recover from transition deviation.

The $8\!\times\!8$ configuration evaluates a qualitatively different challenge: longer planning horizons, where \abbrev is the only method to achieve perfect reward, and does so with zero variance, a result that no baseline approaches.
Despite performing well on the $6\!\times\!6$ map, PPO's performance drops to $0.79$ as the horizon lengthens beyond what its reactive policy can reliably navigate.
SAYCAN's falls to $0.62$, performing worse than PPO alone, suggesting that step-level LM guidance without plan commitment becomes a liability rather than an asset at longer horizons.
ASK's performance reaches $0.74$, confirming that selective querying over PPO is beneficial, but the absence of plan-level commitment prevents it from closing the gap.
The SLM's performance is only $0.42$, despite $100\%$ of its actions being issued through the LM, and with extreme length variance ($35.4 \pm 37.6$), a clear sign of directionless wandering resulting from step-level generation without verification or commitment.
\abbrev's LM usage increases to $81.2\%$ while replanning only $1.06$ times per episode on average, suggesting that a single well-verified plan suffices to navigate the longer horizon reliably.

Taken together, \abbrev's LM usage ($27.9\%$ deterministic, $58.4\%$ stochastic, $81.2\%$ long-horizon) tracks the deliberative demands of each setting rather than being a fixed design choice.
Crucially, across all settings the deciding factor is not how often the LM is consulted, but whether its outputs are structured, verified, and committed to execution.

\section{Related Work}

Integrating LMs into reinforcement learning has been explored as a means of improving generalisation beyond training conditions.
\citet{saycan2022arxiv} propose a system in which LM-proposed actions are scored by a learned affordance function and executed one step at a time, with no commitment to future actions and relying on large models unsuitable for low-latency deployment.
\citet{monteiro2026ask} (ASK) address the computational complexity by selectively querying the LM when the agent's epistemic uncertainty exceeds a certain threshold, improving OOD generalisation without the need for retraining.
However, ASK still treats the LM as a step-level advisor, and thus cannot leverage the structured guidance that committed plans provide, leading to performance degradation as task difficulty increases.

The reliability of LMs as planners, however, remains limited.
\citet{huang2022language} show that while large LMs can propose plausible action sequences from natural language descriptions, their outputs are not grounded in environment dynamics, thus failing in execution.
Recent studies~\cite{ValmeekamKambhampati2025Frontier,armony2025far} further demonstrate that even frontier models struggle to maintain consistency in long-horizon tasks requiring precise state tracking.
\abbrev accounts for these limitations: rather than treating the SLM as an autonomous planner, it wraps plan generation in a simulation-grounded verification loop, ensuring that feasible, complete, and safe plans are committed to execution.

\section{Conclusion and Limitations}

In this work, we proposed \method (\abbrev), a hybrid architecture that combines a fast, reactive RL policy with a slow, deliberative SLM planner through structured plan generation, simulation-grounded verification, and committed execution.
Evaluated on FrozenLake across configurations of increasing difficulty, \abbrev outperforms all baselines while operating with a $2$B-parameter LM, demonstrating that deliberative planning and reactive execution are more powerful in concert than either is alone.
In particular, \abbrev's commitment mechanism provides robustness under stochasticity that step-level integration approaches cannot achieve, and its verification loop relaxes the model-scale requirement for effective integration.

Despite these encouraging results, \abbrev has limitations that motivate future work.
First, the plan generation module relies on a hand-crafted transition function to simulate candidate plans; while this can be replaced by a learned dynamics model~\cite{gavenski2026towards}, the requirement for some form of transition function approximation limits applicability to environments where such a model is available or learnable.
Second, \abbrev has been evaluated exclusively on direct-goal tasks rather than sub-goal decomposition, such as the ``Door Key'' environment from MiniGrid~\cite{towers2024gymnasium}.
However, we hypothesise that \abbrev's architecture is well-suited to such tasks, as the SLM can generate and verify plans that involve sub-goal decomposition, but empirical validation in this setting remains future work.

We believe \abbrev paves the way for a broader research agenda at the intersection of planning and SLMs: how to integrate already trained agents with smaller, more practical LMs without sacrificing performance or costly retraining.
As the community continues to explore what LMs can contribute to planning, we hope \abbrev offers a concrete instantiation of the principle that structured commitment, not scale, is a key ingredient for reliable LM-guided behaviour in goal-directed sequential decision-making.

\section*{Acknowledgment}
This work was partially supported by UK Research and Innovation [grant number EP/S023356/1], in the UKRI Centre for Doctoral Training in Safe and Trusted Artificial Intelligence, and by the Kunumi Institute, through individual grants awarded to the authors.

\clearpage
\bibliography{example_paper}

@article{wang2024comprehensive,
    author = {Wang, Fali and Zhang, Zhiwei and Zhang, Xianren and Wu, Zongyu and Mo, TzuHao and Lu, Qiuhao and Wang, Wanjing and Li, Rui and Xu, Junjie and Tang, Xianfeng and He, Qi and Ma, Yao and Huang, Ming and Wang, Suhang},
    title = {A Comprehensive Survey of Small Language Models in the Era of Large Language Models: Techniques, Enhancements, Applications, Collaboration with LLMs, and Trustworthiness},
    year = {2025},
    issue_date = {December 2025},
    publisher = {Association for Computing Machinery},
    address = {New York, NY, USA},
    volume = {16},
    number = {6},
    issn = {2157-6904},
    url = {https://doi.org/10.1145/3768165},
    doi = {10.1145/3768165},
    journal = {ACM Trans. Intell. Syst. Technol.},
    month = nov,
    articleno = {145},
    numpages = {87}
}

@article{kirk2023survey,
  title={A survey of zero-shot generalisation in deep reinforcement learning},
  author={Kirk, Robert and Zhang, Amy and Grefenstette, Edward and Rockt{\"a}schel, Tim},
  journal={Journal of Artificial Intelligence Research},
  volume={76},
  pages={201--264},
  year={2023}
}

@article{ValmeekamKambhampati2025Frontier,
  title={A Systematic Evaluation of the Planning and Scheduling Abilities of the Reasoning Model o1},
  author={Karthik Valmeekam and Kaya Stechly and Atharva Gundawar and Subbarao Kambhampati},
  journal={Transactions on Machine Learning Research},
  issn={2835-8856},
  year={2025},
  url={https://openreview.net/forum?id=FkKBxp0FhR},
  note={}
}

@article{monteiro2026ask,
  title = "When to ASK: Uncertainty-Gated Language Assistance for Reinforcement Learning",
  author = "Juarez Monteiro and Nathan Gavenski and Gianlucca Zuin and Adriano Veloso",
  year = "2026",
  month = apr,
  day = "1",
  language = "English",
  booktitle = "IEEE World Congress on Computational Intelligence",
  note = "2026 International Joint Conference on Neural Networks (IJCNN), IJCNN ; Conference date: 21-06-2026 Through 26-06-2026",
  url = "https://attend.ieee.org/wcci-2026/",
}

@inbook{gavenski2026towards,
  title = "Towards Generalisable Imitation Learning Through Conditioned Transition Estimation and Online Behaviour Alignment",
  author = "Gavenski, Nathan and Leonetti, Matteo and Rodrigues, Odinaldo",
  year = "2026",
  month = jan,
  day = "21",
  language = "English",
  booktitle = "The 25th International Conference on Autonomous Agents and Multiagent Systems",
  note = "25th International Conference on Autonomous Agents and Multiagent Systems, AAMAS ; Conference date: 25-05-2026 Through 29-05-2026",
  url = "https://cyprusconferences.org/aamas2026/",
}

@misc{armony2025far,
  title = {How far are {LLM}s from symbolic planners? {A}n {NLP}-based perspective},
  author = {Armony, Ma'ayan and Mero{\~n}o-Pe{\~n}uela, Albert and Canal, Gerard},
  year = {2025},
  note = {arXiv:2508.01300}
}

@book{kahneman2012thinking,
  year = {2011},
  author = {Kahneman, Daniel},
  address = {London},
  booktitle = {Thinking, fast and slow},
  isbn = {9780141033570},
  language = {eng},
  publisher = {Penguin},
  title = {Thinking, fast and slow },
}

@inproceedings{saycan2022arxiv,
    title={Do As I Can and Not As I Say: Grounding Language in Robotic Affordances},
    author={Michael Ahn and Anthony Brohan and Noah Brown and Yevgen Chebotar and Omar Cortes and Byron David and Chelsea Finn and Chuyuan Fu and Keerthana Gopalakrishnan and Karol Hausman and Alex Herzog and Daniel Ho and Jasmine Hsu and Julian Ibarz and Brian Ichter and Alex Irpan and Eric Jang and Rosario Jauregui Ruano and Kyle Jeffrey and Sally Jesmonth and Nikhil Joshi and Ryan Julian and Dmitry Kalashnikov and Yuheng Kuang and Kuang-Huei Lee and Sergey Levine and Yao Lu and Linda Luu and Carolina Parada and Peter Pastor and Jornell Quiambao and Kanishka Rao and Jarek Rettinghouse and Diego Reyes and Pierre Sermanet and Nicolas Sievers and Clayton Tan and Alexander Toshev and Vincent Vanhoucke and Fei Xia and Ted Xiao and Peng Xu and Sichun Xu and Mengyuan Yan and Andy Zeng},
    booktitle={arXiv preprint arXiv:2204.01691},
    year={2022}
}

@book{bratman1999intention,
  year = {1999},
  author = {Bratman, Michael},
  address = {Stanford, Calif},
  booktitle = {Intention, plans, and practical reason},
  isbn = {9781575861920},
  publisher = {Center for the Study of Language and Information},
  title = {Intention, plans, and practical reason },
}

@inproceedings{huang2022language,
  title={Language models as zero-shot planners: Extracting actionable knowledge for embodied agents},
  author={Huang, Wenlong and Abbeel, Pieter and Pathak, Deepak and Mordatch, Igor},
  booktitle={International conference on machine learning},
  pages={9118--9147},
  year={2022},
  organization={PMLR}
}

@inproceedings{kendall2017what,
  author = {Alex Kendall and Yarin Gal},
  title = {What uncertainties do we need in {Bayesian} deep learning for computer vision?},
  booktitle = {Advances in Neural Information Processing Systems 30 ({NeurIPS})},
  pages = {5574--5584},
  year = {2017}
}

@article{schulman2017proximal,
  title={Proximal policy optimization algorithms},
  author={Schulman, John and Wolski, Filip and Dhariwal, Prafulla and Radford, Alec and Klimov, Oleg},
  journal={arXiv preprint arXiv:1707.06347},
  year={2017}
}

@article{towers2024gymnasium,
  title={Gymnasium: A standard interface for reinforcement learning environments},
  author={Towers, Mark and Kwiatkowski, Ariel and Terry, Jordan and Balis, John U and De Cola, Gianluca and Deleu, Tristan and Goul{\~a}o, Manuel and Kallinteris, Andreas and Krimmel, Markus and KG, Arjun and others},
  journal={arXiv preprint arXiv:2407.17032},
  year={2024}
}

@misc{qwen3.5,
    title  = {{Qwen3.5}: Towards Native Multimodal Agents},
    author = {{Qwen Team}},
    month  = {February},
    year   = {2026},
    url    = {https://qwen.ai/blog?id=qwen3.5}
}
\bibliographystyle{icml2026}

%%%%%%%%%%%%%%%%%%%%%%%%%%%%%%%%%%%%%%%%%%%%%%%%%%%%%%%%%%%%%%%%%%%%%%%%%%%%%%%
%%%%%%%%%%%%%%%%%%%%%%%%%%%%%%%%%%%%%%%%%%%%%%%%%%%%%%%%%%%%%%%%%%%%%%%%%%%%%%%
% APPENDIX
%%%%%%%%%%%%%%%%%%%%%%%%%%%%%%%%%%%%%%%%%%%%%%%%%%%%%%%%%%%%%%%%%%%%%%%%%%%%%%%
%%%%%%%%%%%%%%%%%%%%%%%%%%%%%%%%%%%%%%%%%%%%%%%%%%%%%%%%%%%%%%%%%%%%%%%%%%%%%%%
\newpage
\appendix
\onecolumn
\section{\method Architecture and Prompts}\label{ap:arch_and_prompts}

"The three Plan, Align, Commit, Think modules share the same SLM backbone, but differ in their prompts and the information they receive from the environment.
Results in Table~\ref{tab:results} show the performance of \abbrev with a Qwen$3.5$--$2$B.
For the RL policy, we use a PPO to train an Actor-Critic architecture presented in Table~\ref{tab:neural_network} with a dropout rate of $0.2$.

\begin{table}[h]
  \scriptsize
  \centering
  \caption{PPO Actor-Critic Network Architecture (FrozenLake $6\!\times\!6$).}
  \label{tab:neural_network}
    \begin{tabular*}{.5\linewidth}{@{\quad}l@{\extracolsep{\fill}}lcc@{\quad}}
      \toprule\noalign{\vspace{-1pt}}
      \textbf{Branch} & \textbf{Layer (Type)} & \textbf{Input} & \textbf{Output} \\[-2pt]
      \midrule
      Input & Flatten Extractor & $72$ & $72$ \\[-2pt]
      \midrule
      \multirow{6}{*}{Policy Net ($\pi$)}
      & Linear         & $72$ & $64$ \\[-2pt]
      & Dropout ($p=0.2$) & $64$ & $64$ \\[-2pt]
      & Tanh           & $64$ & $64$ \\[-2pt]
      & Linear         & $64$  & $64$ \\[-2pt]
      & Dropout ($p=0.2$) & $64$  & $64$ \\[-2pt]
      & Tanh           & $64$  & $64$ \\[-2pt]
      \midrule
      \multirow{6}{*}{Value Net ($vf$)}
      & Linear         & $72$ & $64$ \\[-2pt]
      & Dropout ($p=0.2$) & $64$ & $64$ \\[-2pt]
      & Tanh           & $64$ & $64$ \\[-2pt]
      & Linear         & $64$  & $64$ \\[-2pt]
      & Dropout ($p=0.2$) & $64$  & $64$ \\[-2pt]
      & Tanh           & $64$  & $64$ \\[-2pt]
      \midrule
      Action Head & Linear & $64$ & $4$ \\[-2pt]
      Value Head  & Linear & $64$ & $1$ \\[-2pt]
      \bottomrule
    \end{tabular*}
\end{table}

\subsection{Planner Prompt}
As displayed in Figure~\ref{fig:method}, the planner module receives a task description (Prompt~\ref{lst:prompt_planner} Lines~\ref{prompt:task_start}--\ref{prompt:task_end}) and the observation (Prompt~\ref{lst:prompt_planner} Lines~\ref{prompt:grid_start}--\ref{prompt:grid_end}).
If prompted by the environment, the prompt has the first observation $o_1$, and if by the verification or alignment module, the last observation before breaking any constraints or alignment failure, respectively.
Moreover, the planner module receives information indicating which actions are safe and which are unsafe.
However, it has the freedom to choose any possible actions from the environment (even those that are invalid, such as moving to an edge).

\begin{listing}[h]
\begin{lstlisting}
    <|im_start|>system
        You are a robot navigation agent on a grid. %\label{prompt:task_start}%
        You receive the full grid map, your position, the goal position, and your move history.
        You must choose exactly one action: UP, DOWN, LEFT, or RIGHT.
        You must NEVER choose an action listed under UNSAFE actions.
        Reply with one word only.
        No explanation. %\label{prompt:task_end}%
    <|im_end|>
    <|im_start|>user
        You are a navigation agent on an 8x8 grid.

        Grid (A=Agent, G=Goal, H=Hole, F=Free, S=Start, #=visited multiple times): %\label{prompt:grid_start}%
        A H F F F F F H
        F F F F F F F F
        F F H F F F H F
        F F F F F F F H
        H H H F F F F F
        F F F H F F H H
        F F F F F F F F
        F F F F F F H G %\label{prompt:grid_end}%

        Agent position : row=0, col=0
        Goal position  : row=7, col=7

        SAFE actions   : DOWN
        INVALID actions : UP(edge), LEFT(edge), RIGHT(hole) - do NOT choose these
        This is your first move.
        If there are other moves which are safe but not in your last moves list, prefer these.
        Choose one of the SAFE actions that moves you toward the goal while avoiding holes.
        If no safe action moves closer, choose any safe action to go around obstacles.

        Reply with exactly ONE of the following words: UP, DOWN, LEFT, RIGHT.
    <|im_end|>
\end{lstlisting}
\caption{Prompt for planner module for \abbrev}
\label{lst:prompt_planner}
\end{listing}

We note that during the slipperiness configuration, the hand-crafted transition function simulates a deterministic environment, and thus the planner module receives the same observation for the same action, even if the environment is stochastic.
\abbrev delegates the responsibility of handling stochasticity to the alignment module, which can replan when the environment deviates from the verified plan.

\subsection{Verification Prompt}

The verification module has two options: (i) try to fix the candidate plan by generating a new one based on the feedback from the verification process, or (ii) ask the planner module to generate a new candidate plan providing some feedback.
Prompt~\ref{lst:prompt_verification} shows an example of the prompt for the verification module.
Lines~\ref{prompt:verification_start}--\ref{prompt:verification_end} describe the task and the expected output, while Lines~\ref{prompt:verification_grid_start}--\ref{prompt:verification_grid_end} provide the grid, the candidate plan, and the violations that need to be fixed.

\begin{listing}[h]
\begin{lstlisting}
    <|im_start|>system
        You are a safety verification agent.  %\label{prompt:verification_start}%
        You will be given a navigation plan that has been found to be unsafe, along with a description of the violations.
        Your task is to produce a corrected plan that avoids all Holes and stays within the grid bounds.
        Respond only with the corrected action list, e.g. [RIGHT, DOWN, RIGHT, DOWN].
        Do not explain. %\label{prompt:verification_end}%
    <|im_end|>
    <|im_start|>user
        The following plan is UNSAFE or INCOMPLETE. %\label{prompt:verification_grid_start}%

        Grid:
        S F F F H F F F
        F H F F H F H F
        F H F H F F F F
        F F F F F H H F
        F F H F F H F H
        F F F F F F F F
        F F F F F F F F
        H F F F F H H G

        Plan: ['DOWN', 'DOWN', 'DOWN', 'RIGHT', 'RIGHT', 'DOWN', 'RIGHT', 'RIGHT', 'RIGHT', 'RIGHT', 'RIGHT', 'DOWN']

        Violations:
          - Step 5: action DOWN leads to (4, 2) which is a hazard(H).
        Additionally, the plan does not reach the goal.

        Provide a corrected plan from S to G avoiding all hazard cells (H). Output ONLY a bracket list: [ACTION, ACTION, ...] %\label{prompt:verification_grid_end}%
    <|im_end|>
\end{lstlisting}
\caption{Prompt for verification module for \abbrev}
\label{lst:prompt_verification}
\end{listing}

If it fails to provide a safe plan, it will prompt the planner module with the last observation before the violation and the feedback from the verification process, and ask it to generate a new candidate plan using Prompt~\ref{lst:prompt_planner}.
This process is repeated until both modules agree on a plan that is safe, complete, and feasible.

\subsection{Alignment Prompt}

The alignment module tries to reconcile the agent's current position with the candidate plan since the planner module operates under the belief the agent never moved, and the RL policy operates reactively without awareness of the candidate plan.
Prompt~\ref{lst:prompt_alignment} shows an example of the prompt for the alignment module.
Lines~\ref{prompt:alignment_start}--\ref{prompt:alignment_end} describe the task and the expected output, while Lines~\ref{prompt:alignment_grid_start}--\ref{prompt:alignment_grid_end} provide the grid, the agent's actual position, the expected position, the goal, the safe and unsafe actions, and the intended plan.

\begin{listing}[h]
\begin{lstlisting}
    <|im_start|>system
      You are an alignment agent. %\label{prompt:alignment_start}%
      A navigation agent has diverged from its intended trajectory.
      You will be given the original plan, the agent's current position, and where the agent was expected to be.
      Your task is to recommend the single best next action to safely guide the agent toward the goal, taking both the intended plan and the agent's actual position into account.
      Output format: respond with exactly ONE word. The word must be one of: UP, DOWN, LEFT, RIGHT.
      No other text. No explanation. No punctuation. Just the single action word. %\label{prompt:alignment_end}%
    <|im_end|>
    <|im_start|>user
        Navigation agent needs the single best next action. %\label{prompt:alignment_grid_start}%

        Grid (A=Agent, E=Expected position, G=Goal, H=Hole, S=Start):
        S H A F F F F H
        F F F E F F F F
        F F H F F F H F
        F F F F F F F H
        H H H F F F F F
        F F F H F F H H
        F F F F F F F F
        F F F F F F H G

        Agent actual position  : row=0, col=2
        Expected position      : row=1, col=3
        Goal                   : (7,7)

        SAFE actions   : DOWN, RIGHT
        UNSAFE actions : UP(out of bounds), LEFT(hole) --- do NOT choose these

        Intended plan  : ['DOWN', 'RIGHT', 'RIGHT', 'RIGHT', 'RIGHT', 'RIGHT', 'DOWN', 'DOWN', 'DOWN', 'DOWN', 'DOWN', 'RIGHT', 'RIGHT', 'DOWN']
        Remaining steps: ['RIGHT', 'RIGHT', 'DOWN', 'DOWN', 'DOWN', 'DOWN', 'DOWN', 'RIGHT', 'RIGHT', 'DOWN']

        Choose one of the SAFE actions that moves closer to the goal while staying aligned with the intended plan.
        Reply with exactly ONE word: the action name.  %\label{prompt:alignment_grid_end}%
    <|im_end|>
\end{lstlisting}
\caption{Prompt for alignment module for \abbrev}
\label{lst:prompt_alignment}
\end{listing}

\section{Uncertainty Threshold}\label{ap:threshold}

The epistemic uncertainty threshold $\tau$ controls when the fast component pauses and waits for a plan from the slow component.
We select $\tau$ using a held-out validation set of contexts that is disjoint from both the $100$ training contexts and the test contexts used to report results, avoiding any leakage between threshold selection and final evaluation.
To set $\tau$, we compute the mean epistemic uncertainty $\mathcal{U}(o)$ of the fast component across all observations in this validation set, keeping $\tau$ fixed across all three map configurations.
The average uncertainty was $\approx0.09$, corresponding to observations where the RL policy is operating within its trained distribution.
We set $\tau = 0.1$, slightly above this average, so that the slow component is only invoked when the fast component's uncertainty is above its typical operating level, indicating a genuinely unfamiliar observation rather than normal decision-making noise.

\section{Baseline Adaptations}\label{ap:baselines}

ASK~\cite{monteiro2026ask} was originally evaluated on FrozenLake and is used here without domain adaptation, replacing only the language model backbone with the same Qwen$3.5$--$2$B used by \abbrev.
SAYCAN~\cite{saycan2022arxiv} was adapted to FrozenLake by replacing its robotic affordance scoring with the PPO value head, which scores each LM-proposed action by its estimated future return before execution.

%%%%%%%%%%%%%%%%%%%%%%%%%%%%%%%%%%%%%%%%%%%%%%%%%%%%%%%%%%%%%%%%%%%%%%%%%%%%%%%
%%%%%%%%%%%%%%%%%%%%%%%%%%%%%%%%%%%%%%%%%%%%%%%%%%%%%%%%%%%%%%%%%%%%%%%%%%%%%%%

\end{document}